\documentclass[pdflatex,sn-mathphys-num]{sn-jnl}

\usepackage{graphicx}%
\usepackage{multirow}%
\usepackage{amsmath,amssymb,amsfonts}%
\usepackage{amsthm}%
\usepackage{mathrsfs}%
\usepackage[title]{appendix}%
\usepackage{xcolor}%
\usepackage{textcomp}%
\usepackage{manyfoot}%
\usepackage{booktabs}%
\usepackage{algorithm}%
\usepackage{algorithmicx}%
\usepackage{algpseudocode}%
\usepackage{listings}%
\usepackage{lineno}

\usepackage{float}
\usepackage{caption}
\usepackage{multirow}

\theoremstyle{thmstyleone}%
\theoremstyle{thmstyletwo}%

\theoremstyle{thmstylethree}%

\newcommand{\figref}[1]{Fig.\ref{#1}}

\newcommand{\secref}[1]{Sec.\ref{#1}}

\newcommand{\eg}{\textit{e}.\textit{g}.}

\newcommand{\hbz}[1]{{ {#1}}}

\newcommand{\hl}[1]{{ {#1}}}

\begin{document}

\title[Feed-Forward Multi-View Multi-Person Reconstruction with Contrastive Human-Aware 3D Representation]{Feed-Forward Multi-View Multi-Person Reconstruction with Contrastive Human-Aware 3D Representation}


\author[1]{\fnm{Yuanwang} \sur{Yang}}\equalcont{Equal contribution.}\email{yyw@tju.edu.cn} 

\author[1]{\fnm{Buzhen} \sur{Huang}}\equalcont{Equal contribution.}\email{hbz@tju.edu.cn}

\author[1]{\fnm{Zongxuan} \sur{Ren}}\email{xuan591847902@163.com}

\author[1]{\fnm{Jing} \sur{Huang}}\email{hj00@tju.edu.cn}

\author*[1]{\fnm{Kun} \sur{Li}}\email{lik@tju.edu.cn}

\affil[1]{\orgdiv{College of Intelligence and Computing}, \orgname{Tianjin University}, \orgaddress{\street{135 Yaguan Road}, \city{Tianjin}, \postcode{300350}, \state{Tianjin}, \country{China}}} 
\abstract{\hbz{Multi-view human reconstruction has been extensively studied under simplified settings, yet scaling these methods to robust and efficient multi-person reconstruction in unconstrained environments requires a more general and scalable modeling paradigm. Existing bottom-up methods often rely on accurate camera calibration and explicit cross-view matching, and therefore struggle in multi-person scenarios with severe occlusions and ambiguities. We attribute these limitations to the lack of an explicit and view-agnostic 3D representation for jointly reasoning about human structure and multi-view consistency. Based on this observation, we propose a new top-down paradigm that maintains a unified, instance-centric human-aware 3D space, enabling simultaneous camera calibration, cross-view association, and human reconstruction via cross-modal contrastive learning. Specifically, observations from multiple views are lifted and fused directly into this shared 3D space, where geometric structure, visual appearance, and human-centric semantic cues are jointly encoded at the instance level. To enhance the discriminability and consistency of the 3D representation, we introduce a spatial contrastive learning strategy that aligns 3D features corresponding to the same human instance across different views and modalities, while separating features from different instances. This design allows correspondence reasoning, semantic aggregation, and instance discrimination to be performed natively in the 3D space, which enforces cross-view consistency and improves robustness under severe occlusions. Based on the learned instance-aware 3D representation, we recover structured human body models in a feed-forward manner by regressing SMPL parameters from instance-level 3D human tokens. Extensive experiments demonstrate that adopting a unified human-aware 3D feature space as the core representation leads to robust, accurate, and efficient multi-view human reconstruction in challenging real-world scenarios. The code and data will be made publicly available.}

}

\keywords{Unconstrained 3D reconstruction, Multi-person pose and shape estimation, Human-scene interaction}

\maketitle

\begin{center}
    \captionsetup{type=figure}
    \includegraphics[width=\textwidth]{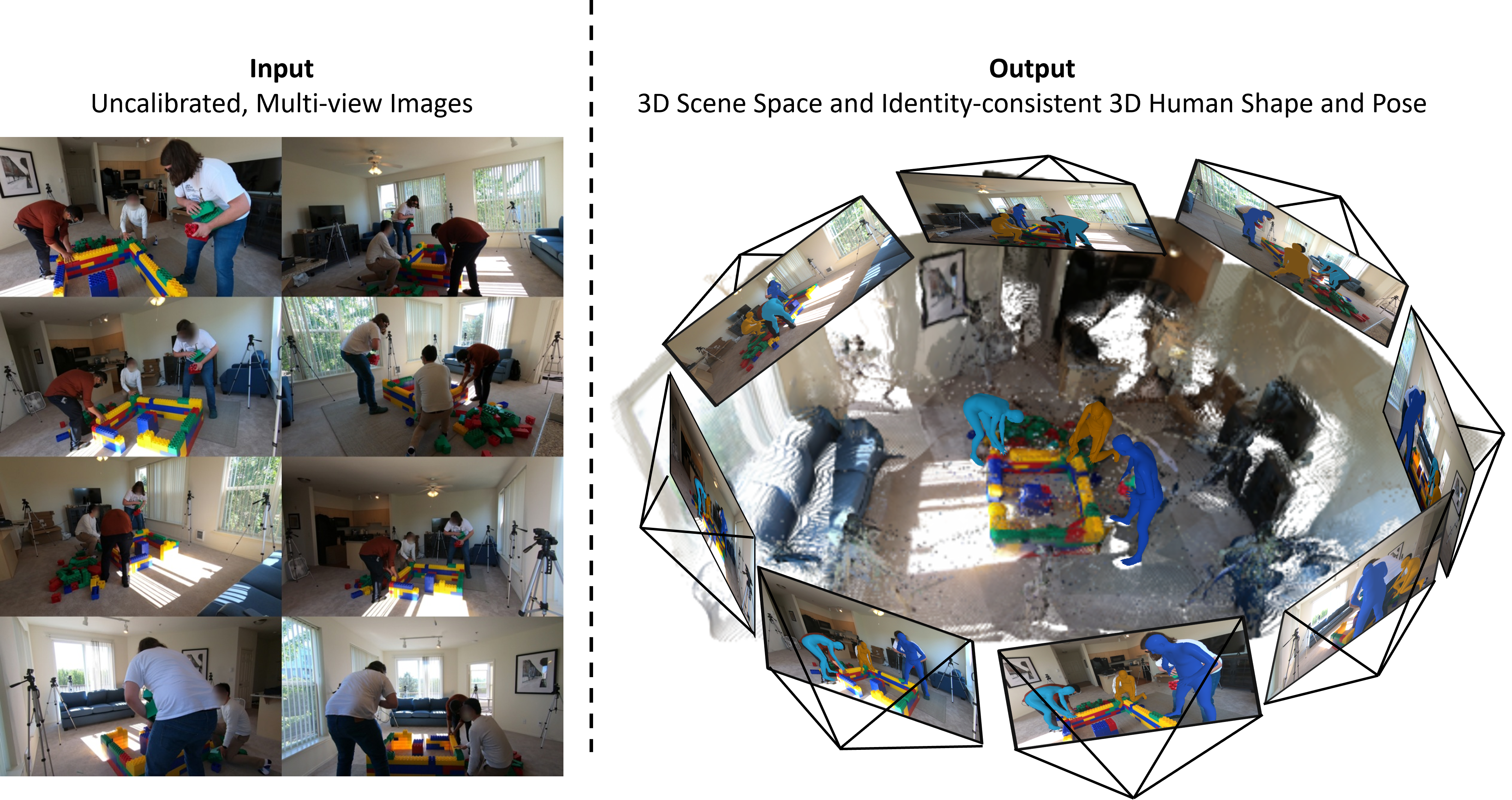}
    \captionof{figure}{We propose a unified, human-aware 3D representation learned via cross-modal contrastive learning. This representation provides a persistent and view-agnostic 3D space in which camera calibration, cross-view association, and human reconstruction can be jointly performed.}
    \label{teaser}
\end{center}
\section{Introduction}
\label{sec:intro}
\hbz{
Reconstructing 3D humans from multiple camera views is a long-standing problem with wide applications in VR/AR, sports analysis, and human–computer interaction. Although extensive research has been devoted to this topic, achieving robust and efficient reconstruction in complex real-world scenarios remains challenging. Traditional multi-view systems~\cite{joo2015panoptic,ionescu2013human3,dong2021fast,tu2020voxelpose} rely heavily on accurate camera calibration and well-controlled environments, making them fragile under practical conditions such as imperfect setups, heavy occlusions, or large-scale scenes (\figref{sec:intro}). 

The majority of existing multi-view human reconstruction approaches~\cite{dong2021fast,zhang2021direct,huang2023simultaneously} follow a bottom-up “2D-to-3D” paradigm.
They first extract 2D cues~(\eg, keypoints, silhouettes, or per-view body parameters) from each image, establish cross-view correspondences, and then fuse these observations into a 3D representation. As one of the dominant approaches, triangulation-based methods directly associate and fuse multi-view 2D observations with camera geometry, but minor inaccuracies in camera parameters can propagate and result in global geometric inconsistencies~\cite{dong2021fast,zhang2021direct,huang2020end,tu2020voxelpose}. Although some works~\cite{huang2023simultaneously,yu2022multiview,huang2023dmmr,muller2025reconstructing} relax the reliance on precise camera geometry with semantic association, they remain inherently vulnerable to occlusion, as missing or corrupted 2D observations directly undermine correspondence estimation and subsequent feature fusion. In addition, these methods often depend on iterative optimization or matching procedures, which are computationally expensive and difficult to scale to complex scenes or real-time applications.


These challenges are not merely implementation issues, but are inherent to the bottom-up paradigm, which reasons about 3D structure through fragmented and view-dependent 2D observations.
In contrast, humans and objects exist and interact in a shared 3D space, and a top-down approach enforces multi-view consistency by reasoning directly within a unified 3D representation rather than across independent image views.
This motivates a shift in perspective: instead of treating 3D reconstruction as the result of fusing 2D predictions, we advocate for maintaining an explicit, shared 3D representation as the central reasoning space throughout the reconstruction process.

In this work, we propose a new top-down paradigm that maintains a unified human-aware 3D space to enable simultaneous camera calibration, cross-view association, and human reconstruction. With this 3D space, multi-view information is lifted and fused at the feature level, where geometric structure, appearance cues, and human-centric semantics are jointly encoded in a shared representation. To enhance the discriminability and consistency of the 3D representation, a spatial contrastive learning strategy is further introduced. Consequently, this 3D feature space serves as a persistent, view-agnostic set of spatial tokens, and thus cross-view association, occlusion handling, and instance reasoning can be performed natively within the 3D space.


Specifically, given uncalibrated multi-view images, we initialize the 3D space with a geometry-aware prior~\cite{wang2025vggt} and subsequently finetune the representation through differentiable 3D Gaussian Splatting (3DGS) rendering~\cite{kerbl20233d}. After initialization, the 3D representation encodes only limited human prior knowledge and lacks explicit instance discriminability. To this end, we further introduce a spatial contrastive learning strategy to improve the consistency and discriminability of the 3D feature space. In particular, human points in the 3D space are re-projected into pose, appearance, and geometry fields to sample expressive human semantics. We then apply contrastive learning across sampled features from different views and instances to enforce semantic consistency. Meanwhile, the sampled features are used to regress identity embeddings, which are further contrasted across the 3D and 2D spaces. Spatial contrastive learning not only enriches the human prior knowledge within the 3D space, but also improves geometric alignment~(\eg, camera parameters) by enforcing semantic consistency. With the learned 3D representation, structured SMPL parameters~\cite{loper2023smpl} are finally regressed in a feed-forward manner from instance-aware 3D human tokens.



In summary, our main contributions are as follows:
\begin{itemize}
    \item We propose a novel multi-view human reconstruction framework that explicitly maintains a unified 3D feature space as the central representation, which enables feed-forward and simultaneous calibration, association and reconstruction.
    \item We design a human-aware 3D representation and refinement strategy that integrates multi-view geometry, appearance, and human semantics at the representation level, thereby substantially improving robustness to occlusion and camera calibration errors.
    \item We introduce a spatial contrastive learning strategy that leverages semantic consistency to enhance geometric alignment and reconstruction accuracy in multi-view, multi-person scenarios.
    
\end{itemize}

}
\section{Related Work}
\label{sec:Related}

\subsection{Human Reconstruction from Unconstrained Cameras}
Real-world deployments often involve unknown, unsynchronized, or distorted cameras, as well as sparse and wide-baseline viewpoints. The key challenge is that uncalibrated cameras prevent direct enforcement of multi-view geometric consistency, so both camera parameters and 3D humans must be inferred from weak and noisy cues; this coupling is particularly brittle under occlusion, wide baselines, distortion, and unsynchronized viewpoints. To deal with these unconstrained settings, several methods~\cite{xu2022multi, yu2022multiview, muller2025reconstructing, li2024multi, javerliat2025kineo} jointly estimate camera parameters and 3D human poses. HSfM~\cite{muller2025reconstructing} uses SMPL-derived anthropometric constraints as metric references and jointly optimizes scene structure, human meshes, and camera parameters through bundle adjustment, addressing both scale ambiguity and missing calibration. \hl{Yu et al.}~\cite{yu2022multiview} \hl{map per-view image features to a canonical human surface that is invariant to viewpoint, pose and fuse them via confidence-weighted attention, enabling reconstruction without camera calibration.} Additional approaches~\cite{li2024multi} incorporate depth or 3D re-identification cues for cross-view association but require extra sensors. \hl{EasyRet3D}~\cite{yin2025easyret3d} \hl{addresses uncalibrated multi-view multi-human reconstruction by explicitly modeling cross-view identity association through a tracking-based formulation. However, it introduces a dependency on explicit tracking quality and can be sensitive to association errors in crowded or heavily occluded scenes.} Kineo~\cite{javerliat2025kineo} focuses on self-calibration and geometric recovery by jointly estimating camera intrinsics/extrinsics and 3D structure from weak observations, leveraging robust sampling, temporal consistency, and anthropometric priors to resolve scale and depth ambiguities.
Although these methods demonstrate the feasibility of calibration-free reconstruction, they still rely on accurate 2D detections, robust cross-view matching, or reliable dense semantic correspondences.
\subsection{Multi-View 3D Human Pose and Shape Estimation.}

As a foundational component of 3D human reconstruction, estimating pose from multiple views has been extensively studied~\cite{chen20243d}. Early approaches~\cite{chen20243d} follow a bottom-up pipeline: 2D poses are detected independently per view and subsequently triangulated or fitted by kinematic models to recover 3D joint locations. Such pipelines are highly sensitive to 2D detection noise and often fail under heavy occlusion or multi-person interaction. To mitigate errors introduced during 2D association, voxel-based methods~\cite{tu2020voxelpose,luo2025voxel} aggregate multi-view features into a unified 3D space and perform inference directly on volumetric representations, achieving improved robustness at the cost of substantial computation and memory. Transformer-based methods further enable end-to-end multi-view reasoning. For instance, MvP~\cite{zhang2021direct} leverages learnable joint queries and projective attention to fuse cross-view geometry, while MV-SSM~\cite{chharia2025mv} introduces state-space modeling to enhance generalization to unseen camera layouts. Meanwhile, multi-view human mesh reconstruction has been widely studied, particularly in controlled environments where intrinsic and extrinsic camera parameters are known~\cite{furukawa2009accurate,hu2023sherf,starck2007surface}. Classical optimization-based methods extend single-view body models (e.g., SMPL/SMPL-X~\cite{loper2023smpl}) by jointly refining camera poses, keypoints, silhouettes, and mesh parameters, achieving metrically consistent meshes when calibration and initialization are reliable~\cite{bogo2016keep,li20213d}. More recent learning-based approaches aggregate multi-view features to directly regress a coherent mesh~\cite{baradel2024multi}. 

\subsection{Feed-forward 3D Reconstruction for Scene and Humans.}

Feed-forward 3D reconstruction~\cite{elflein2025light3r,wang2025vggt,yang2025fast3r,cabon2025must3r,jang2025pow3r,li2025mono3r,liu2025regist3r} has recently emerged as an efficient method for predicting dense 3D structures from images directly, without relying on iterative optimization. VGGT introduces a feed-forward multi-view geometry framework that directly predicts camera parameters together with dense geometric outputs (e.g., depth and point-based 3D representations) from one to many images in a single forward pass, avoiding expensive test-time optimization. Building on this capability, recent works increasingly treat VGGT-style predictors as a front-end to bootstrap geometry and camera estimates, and then plug these initialization signals into downstream tasks such as rendering, tracking, interaction understanding, or task-specific refinement.
Meanwhile, recent work has begun to reconstruct humans, cameras, and the surrounding scene in a shared metric coordinate system, rather than treating human mesh recovery and scene reconstruction as separate problems. HSfM~\cite{muller2025reconstructing} tackles this problem from sparse, uncalibrated multi-view images by jointly reconstructing human meshes, the surrounding scene, and cameras, using human observations to stabilize camera estimation and scene geometry. SynCHMR~\cite{Zhao_2024_CVPR} addresses monocular videos by combining human motion recovery with a SLAM-style formulation to estimate metric camera motion, scene geometry, and a consistent human trajectory in a unified coordinate system. JOSH~\cite{liu2025joint} performs optimization-based 4D human--scene reconstruction from monocular videos by leveraging human--scene contact constraints to jointly refine human motion, camera poses, and dense scene geometry, while its distilled variant JOSH3R~\cite{liu2025joint} enables efficient feed-forward inference trained from JOSH pseudo-labels. Human3R~\cite{chen2025human3r} builds on CUT3R~\cite{wang2025continuous} and performs feed-forward human--scene reconstruction from monocular videos, jointly predicting multi-person SMPL-X bodies, camera motion, and dense scene geometry in a shared world coordinate system.Recently, HAMSt3R~\cite{rojas2025hamst3r} extends MASt3R~\cite{leroy2024mast3r} with human-aware heads and performs fully feed-forward reconstruction, and it predicts dense 3D point maps directly from sparse views, enabling efficient joint recovery of the scene and people. Successfully combines feed-forward 3D reconstruction with joint optimization of the scene and human bodies.
In contrast, we construct a unified implicit 3D embedding space that jointly models pose, appearance, and spatial relationships in an end-to-end manner. This enables direct cross-view correspondence learning and feature fusion in 3D space, substantially improving the robustness of multi-human SMPL reconstruction under unconstrained settings. 

\section{Method}
\label{sec:method}

\begin{figure*}[!t]
\centering
\includegraphics[width=\linewidth]{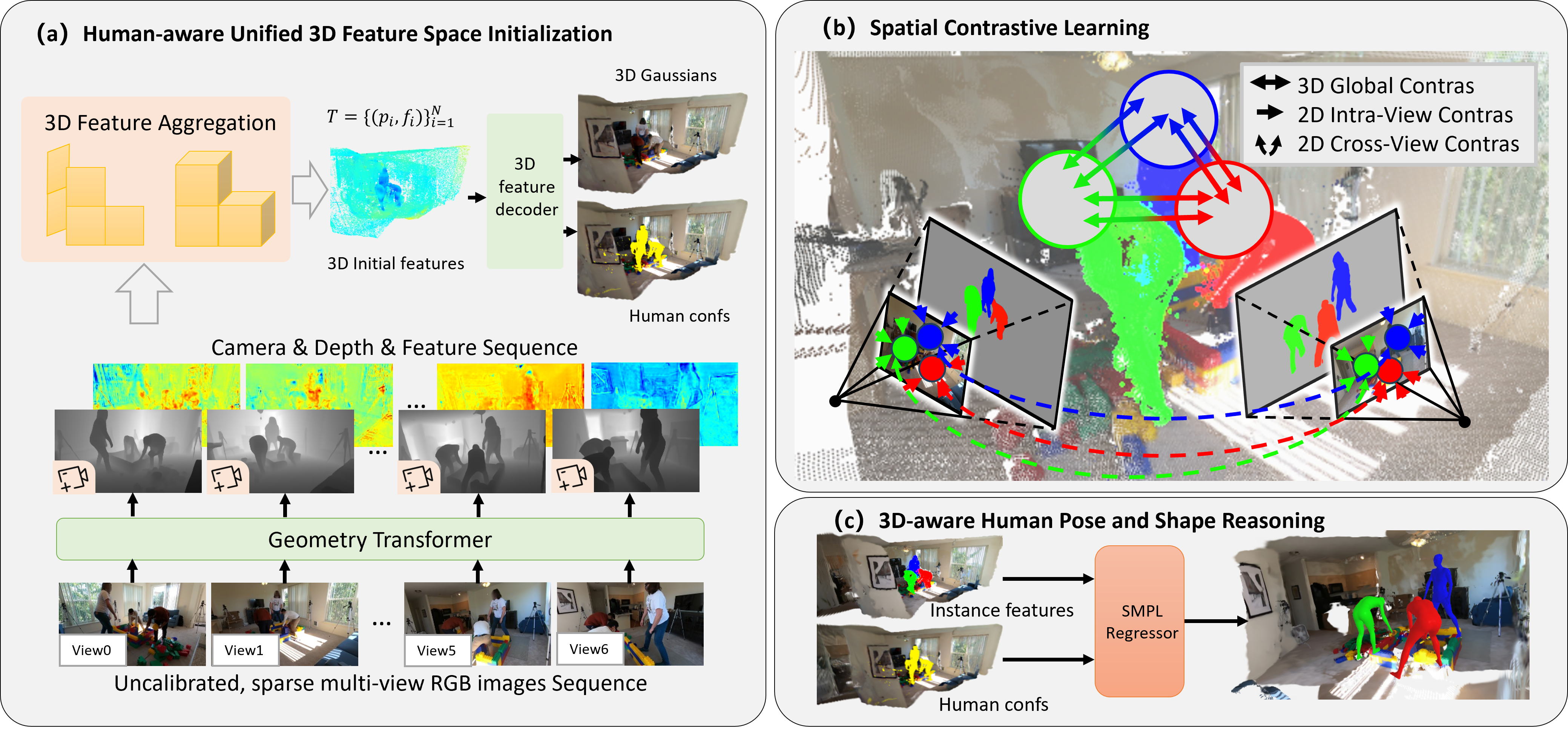}
\caption{\textbf{Overview}. Given a set of multi-view unconstrained images, our method first initializes a 3D space using a geometry-aware prior and 3D Gaussian Splatting~(a). To further enhance human prior knowledge and instance-level discriminability for the 3D representation, spatial contrastive learning is applied by sampling across different semantic fields~(b). Finally, camera and human representations are jointly optimized within the 3D space and regressed in a feed-forward fashion~(c).}
\label{fig:pipeline}
\end{figure*}
\hbz{
As shown in \figref{fig:pipeline}, we present a human-aware multi-view 3D reconstruction framework that explicitly maintains and optimizes a unified 3D feature space as an intermediate representation. Instead of reasoning directly in the 2D image domain or regressing parametric body models from per-view features, our approach encodes geometric structure, semantic information, and human identity within this shared 3D space. This design allows camera calibration, cross-view association, and human reconstruction to be addressed jointly in a feed-forward pipeline. Details of the proposed framework are described in the following sections.



\subsection{Human-aware Unified 3D Feature Space Initialization}
\label{sec:3d_feature_space}

Directly operating on 2D features is inherently fragile under occlusions, viewpoint changes, and inter-view misalignment.
While multi-view aggregation alleviates some of these issues, ambiguity remains when reasoning purely in image space.
We therefore lift image features into 3D and maintain an explicit spatial representation, allowing information to be aggregated, filtered, and refined in a geometry-aware manner, which naturally mitigates these issues.

\subsubsection{Construction of 3D Feature Space.}
As shown in \figref{fig:pipeline}~(a), given a set of uncalibrated images $\{I_v\}_{v=1}^{V}$ from $V$ views, we first construct a view-consistent 3D feature tokens:
\begin{equation}
\mathcal{T} = \left\{ (\mathbf{p}_i, \mathbf{f}_i) \right\}_{i=1}^{N},
\end{equation}
where $\mathbf{p}_i \in \mathbb{R}^3$ denotes a 3D location and $\mathbf{f}_i \in \mathbb{R}^C$ represents a learned 3D feature embedding. $N$ is the number of points and varies with the complexity of the scene.

To initialize the 3D feature space, we employ a pretrained VGGT model~\cite{wang2025vggt} as a geometry-aware prior.
Given each input image $I_v$, VGGT predicts a depth map $D_v$, a per-pixel confidence map $C_v$, a dense feature map $F_v$ and per-view camera parameters $p_v$:
\begin{equation}
(D_v, C_v, F_v, p_v) = \Phi_{\text{VGGT}}(I_v).
\end{equation}

Using the predicted depth and camera parameters, each pixel $\mathbf{u} \in \mathbb{R}^2$ is unprojected into 3D space:
\begin{equation}
\mathbf{x}_v(\mathbf{u}) = \pi^{-1}(\mathbf{u}, D_v(\mathbf{u}); p_v).
\end{equation}
Thus, the image features $F_v(\mathbf{u})$ and corresponding confidence weights $C_v(\mathbf{u})$ can be mapped into 3D space.
\begin{equation}
\mathbf{f}_v(\mathbf{u}) = F_v(\mathbf{u}),
\end{equation}
\begin{equation}
c_v(\mathbf{u}) = C_v(\mathbf{u}).
\end{equation}
This process yields a dense, view-dependent 3D representation, but directly operating on it is computationally expensive and prone to errors from \hl{noisy depth estimation and redundant observations. To obtain a more stable and compact representation, we aggregate neighboring 3D points from multiple views using a confidence-weighted soft aggregation scheme, where aggregation weights are computed as}
\begin{equation}
w_{ij} =  c_i \cdot c_j \cdot \exp\Big(-\frac{\|\mathbf{x}_i - \mathbf{x}_j\|^2}{2\sigma^2}\Big),
\end{equation}
\hl{where $\sigma$ is a fixed hyper-parameter controlling the spatial aggregation range,} and the fused point center $\mathbf{p}_i$ and feature $\mathbf{f}_i$ are then defined as
\begin{equation}
\mathbf{p}_i = \frac{\sum_j \tilde{w}_{ij} \, \mathbf{x}_j}{\sum_j \tilde{w}_{ij}}, \quad
\mathbf{f}_i = \frac{\sum_j \tilde{w}_{ij} \, \mathbf{f}_j}{\sum_j \tilde{w}_{ij}}.
\end{equation}

In contrast to voxelization~\cite{zhang2022voxeltrack,luo2025voxel}, \hl{our representation operates directly on continuous 3D points using soft spatial aggregation, avoiding explicit discretization into fixed voxel grids. This formulation is more flexible with respect to varying scene scales and irregular point distributions.} With the aggregation, the 3D feature space is view-agnostic and spatially anchored, forming a robust 3D feature tokens that integrate multi-view evidence while suppressing noisy or inconsistent observations.





\subsubsection{Human-aware 3D Feature Decoding.}
Since the 3D features from VGGT only describe geometry cues, we then explicitly encode human-related semantics into the 3D feature space.
Each fused 3D feature token $(\mathbf{p}_i, \mathbf{f}_i)$ serves as a shared latent representation from which multiple attributes are decoded.
Specifically, a lightweight MLP $g_{\theta}$ maps $\mathbf{f}_i$ to Gaussian attributes~\cite{kerbl20233d}:
\begin{equation}
(\rho_i, \mathbf{s}_i, \mathbf{q}_i, \mathbf{h}^{\text{SH}}_i, z_i) = g_{\theta}(\mathbf{f}_i),
\end{equation}
where $\rho_i$ denotes density, $\mathbf{s}_i$ scale, $\mathbf{q}_i$ rotation quaternion, $\mathbf{h}^{\text{SH}}_i$ spherical harmonic coefficients, and $z_i$ is a human-related logit.

The Gaussian opacity is obtained via a bounded monotonic mapping:
\begin{equation}
\alpha_i = \psi(\rho_i),
\end{equation}
while the human confidence is defined as
\begin{equation}
h_i = \sigma(z_i).
\end{equation}

The human confidence is defined at the 3D level rather than per image.
This enables consistent human reasoning across views and prevents conflicting predictions caused by occlusion or partial visibility in individual images.
Moreover, both geometric, appearance, and semantic attributes are decoded from the same 3D feature space.
This allows the representation to jointly encode scene structure and human-related semantics, rather than treating human segmentation as a separate 2D task.

\subsubsection{Initial 3D Space Training.}
We train the initial 3D space by learning Gaussian parameters and the human confidence field, without introducing instance-level supervision or SMPL regression.


Specifically, each 3D token is associated with a Gaussian primitive parameterized by its position $\mathbf{p}_i$, scale $\mathbf{s}_i$, rotation $\mathbf{q}_i$, opacity $\alpha_i$, and spherical harmonic coefficients $\mathbf{h}_i^{\text{SH}}$.
Given a camera view $v$, we render color images by splatting these Gaussians onto the image plane.
For a pixel $\mathbf{x}$, the rendered color is computed as
\begin{equation}
\hat{\mathbf{I}}_v(\mathbf{x}) =
\sum_{i}
\mathbf{c}_i(\mathbf{x})
\alpha_i \mathcal{G}_i(\mathbf{x})
\prod_{j<i} \big(1-\alpha_j \mathcal{G}_j(\mathbf{x})\big),
\end{equation}
where $\mathcal{G}_i(\mathbf{x})$ denotes the projected Gaussian footprint and $\mathbf{c}_i(\mathbf{x})$ is the view-dependent color obtained from spherical harmonics.
The photometric reconstruction loss $\mathcal{L}_{\text{rgb}}$ supervises Gaussian appearance by comparing rendered images with the input views:
\begin{equation}
\mathcal{L}_{\text{rgb}} =
\sum_{v}\sum_{\mathbf{x}}
\big\|
\hat{\mathbf{I}}_v(\mathbf{x}) - \mathbf{I}_v(\mathbf{x})
\big\|_1,
\end{equation}
where $\hat{\mathbf{I}}_v$ denotes the image rendered from Gaussians in view $v$.

To supervise human confidence in 3D, we render a soft human mask by modulating Gaussian opacity with the predicted confidence:
\begin{equation}
\tilde{\alpha}_i = \alpha_i \cdot h_i .
\end{equation}
Rasterizing $\tilde{\alpha}_i$ yields a per-view human mask $\hat{M}_v$, which is supervised by available 2D masks $M_v$ using a binary cross-entropy loss:
\begin{equation}
\mathcal{L}_{\text{mask}} =
\sum_v \text{BCE}(\hat{M}_v, M_v).
\end{equation}

We further introduce a regularization term $\mathcal{L}_{\text{reg}}$ to stabilize Gaussian optimization and encourage spatially coherent human confidence.
Specifically, we penalize excessively opaque Gaussians, regularize Gaussian scales to avoid degenerate footprints, and enforce local smoothness of the 3D human confidence field:
\begin{equation}
\mathcal{L}_{\text{reg}}
=
\lambda_{\alpha}\sum_i \alpha_i^2
+
\lambda_{s}\sum_i \|\log \mathbf{s}_i\|_1
+
\lambda_{h}\sum_{i}\sum_{j\in\mathcal{N}(i)} (h_i-h_j)^2,
\label{eq:lreg}
\end{equation}
where $\mathcal{N}(i)$ denotes the spatial neighbors of token $i$ (e.g., within a fixed radius in 3D).





\subsection{3D Feature Enhancement with Spatial Contrastive Learning}
\label{sec:instance_embedding}

The 3D feature space in \secref{sec:3d_feature_space} provides stable anchors but may lose fine-grained details and lacks explicit instance discriminability.
We therefore \emph{enhance} the 3D token features and \emph{learn} instance-discriminative fields with spatial contrastive learning by selectively sampling from more human semantics.
Importantly, both enhancement and instance reasoning are defined on the shared 3D token field, which mitigates view-dependent ambiguities from occlusion and truncation.

\begin{figure*}[!t]
\centering
\includegraphics[width=\linewidth]{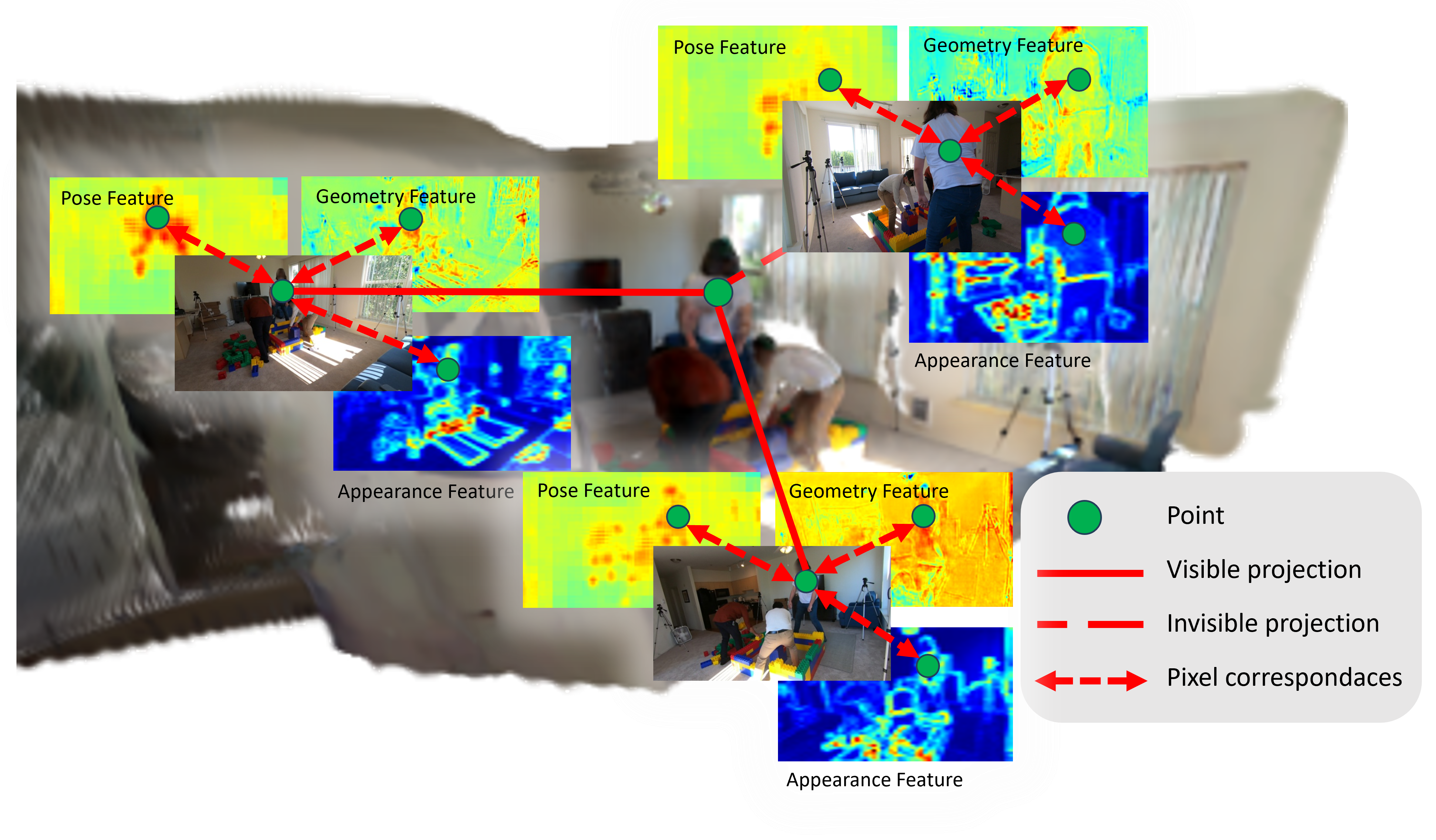}
\caption{Human points in the 3D space are projected into different semantic fields to obtain features for spatial contrastive learning.}
\label{fig:sampling}
\end{figure*}
\subsubsection{Human Semantics Sampling}

After initializing the 3D space, the resulting 3D features contain only limited human-related information; therefore, we further exploit richer semantic cues from the image domain. Specifically, we adopt ROMP~\cite{sun2021monocular} features $P_v$ to provide additional pose information, since they encode pixel-wise human pose and body structure cues. The appearance cues from DINOv3~\cite{simeoni2025dinov3} $A_v$ are also utilized. In addition, since the aggregation adjust the point positions, we further sampling from the VGGT feature maps $F_v$.

However, not all 3D tokens require semantic enhancement. We only focus on the tokens with high human confidence $h_i$, which are more likely to lie on human surfaces:
\begin{equation}
\mathcal{H} = \{ i \mid h_i > \tau_h \},
\end{equation}
where $\tau_h$ is a confidence threshold. Restricting the sampling points to $\mathcal{H}$ significantly reduces computational cost and prevents background clutter from contaminating the refinement process.
Tokens outside $\mathcal{H}$ retain their original geometric features and continue to contribute to scene reconstruction.




For each selected 3D token $(\mathbf{p}_i, \mathbf{f}_i)$ and each camera view $v$, we project the 3D position $\mathbf{p}_i$ onto the image plane:
\begin{equation}
\mathbf{u}_{i}^{(v)} = \pi(\mathbf{p}_i; p_v),
\end{equation}
where $\pi(\cdot)$ denotes the camera projection function defined by the estimated camera parameters.

To ensure reliable feature sampling, we perform visibility-aware filtering.
A reprojection is considered valid if it satisfies two conditions:
(i) the projected location lies within image boundaries, and
(ii) the projected depth is consistent with the predicted depth map.

Formally, the visibility indicator $m_i^{(v)}$ is defined as
\begin{equation}
m_i^{(v)} =
\mathbb{I}(\mathbf{u}_{i}^{(v)} \in \Omega)
\cdot
\mathbb{I}\big(|z_i^{(v)} - D_v(\mathbf{u}_{i}^{(v)})| < \epsilon\big),
\end{equation}
where $\Omega$ denotes the image domain, $z_i^{(v)}$ is the projected depth of $\mathbf{p}_i$ in view $v$, $D_v(\cdot)$ is the predicted depth map, and $\epsilon$ is a tolerance threshold.

As shown in \figref{fig:sampling}, by sampling different semantic maps $P_v$, $A_v$, and $F_v$ with each valid reprojection point $\mathbf{u}_{i}$, we then obtain pose, appearance and geometry features $P_v(\mathbf{u}_{i})$, $A_v(\mathbf{u}_{i})$, and $F_v(\mathbf{u}_{i})$.
}


\subsubsection{Cross-modal Feature Fusion}
\label{sec:cross_modal_fusion}
The sampled features $P_v(\mathbf{u}_{i})$, $A_v(\mathbf{u}_{i})$, and $F_v(\mathbf{u}_{i})$ are concatenated as $\mathbf{g}_{i}^{(v)}$ and then aggregated across multiple views to refine the original 3D feature.
Each view contributes with a confidence weight proportional to its visibility:
\begin{equation}
w_i^{(v)} = \frac{m_i^{(v)}}{\sum_{v'} m_i^{(v')} + \delta},
\end{equation}
where $\delta$ is a small constant for numerical stability.

The refined 3D feature is computed via a residual fusion formulation:
\begin{equation}
\tilde{\mathbf{f}}_i
=
\mathbf{f}_i
+
\sum_{v}
w_i^{(v)} \cdot
\phi\big([\mathbf{f}_i, \mathbf{g}_{i}^{(v)}]\big),
\end{equation}
where $\phi(\cdot)$ is a lightweight fusion network implemented as an MLP, and $[\cdot]$ denotes feature concatenation.

This residual design preserves the stability of the original 3D representation while selectively injecting high-level semantic information.
Importantly, fusion is performed independently for each 3D token, enabling scalable and parallel refinement.

By selectively sampling and fusing 2D features through reprojection, we compensate for the insufficient human information and enable the refined 3D feature space to better align with articulated human structure.
Moreover, repeated reprojection and semantic supervision implicitly enforce cross-view consistency, allowing the method to naturally scale to higher-resolution features and additional modalities without modifying the core architecture.

\subsubsection{3D Instance Embedding and Spatial Contrastive Learning}
\label{sec:instance_embedding}

Based on the refined 3D features, we further learn instance-aware representations to distinguish multiple humans in the shared 3D space.
For each 3D token, we predict a compact instance embedding using a lightweight projection head:
\begin{equation}
\mathbf{e}_i = \frac{h_{\phi}(\tilde{\mathbf{f}}_i)}{\|h_{\phi}(\tilde{\mathbf{f}}_i)\|_2},
\end{equation}
where $h_{\phi}$ denotes an MLP and $\ell_2$ normalization enforces a bounded embedding space.
\hl{The projection head $h_{\phi}(\cdot)$ is optimized during contrastive learning, while the backbone features $\mathbf{f}_i$ remain fixed.}
Instance embeddings are only computed for tokens with high human confidence, which suppresses background interference and reduces unnecessary computation.

To learn discriminative embeddings, we adopt a spatial contrastive learning objective that links 3D tokens with 2D instance supervision.
Each 3D token is associated with a Gaussian primitive parameterized by its fixed geometric attributes (position, scale, rotation, and opacity) and its instance embedding.
During rendering, the instance embedding is treated as the feature carried by the Gaussian and projected onto the image plane via Gaussian splatting.
Geometric parameters are detached during this process to prevent degenerate solutions.

Formally, the rendered instance embedding map at pixel $\mathbf{x}$ is given by
\begin{equation}
\mathbf{E}(\mathbf{x}) =
\sum_{i} \mathbf{e}_i \, \alpha_i \mathcal{G}_i(\mathbf{x})
\prod_{j<i} \big(1-\alpha_j \mathcal{G}_j(\mathbf{x})\big),
\end{equation}
which softly aggregates contributions from visible 3D tokens along each viewing ray.
This differentiable rendering establishes a correspondence between 3D tokens and 2D pixels, allowing instance supervision to be propagated from image space back to the shared 3D representation.

\hl{We apply the same contrastive objective under three complementary supervision domains: intra-view, cross-view, and 3D-space contrastive learning. Intra-view contrastive learning is performed within each image to improve local instance discrimination and suppress ambiguity caused by overlapping humans in image space. Cross-view contrastive learning aligns embeddings belonging to the same person across different camera views, improving view consistency and cross-view association robustness. Finally, 3D-space contrastive learning directly regularizes the shared 3D representation after multi-view feature aggregation, enforcing identity consistency in the unified 3D space. These three forms of supervision are complementary: intra-view supervision improves local separability, cross-view supervision improves identity consistency across viewpoints, while 3D-space supervision stabilizes the globally aggregated instance-aware representation.}
In all cases, embeddings belonging to the same person are encouraged to be close, while embeddings from different persons are separated.
The contrastive loss is defined as
\begin{equation}
\mathcal{L}_{\text{sc}} = -\sum_{i}
\log
\frac{\exp(\mathbf{e}_i^\top \mathbf{u}_{y_i})}
{\sum_{k} \exp(\mathbf{e}_i^\top \mathbf{u}_{y_k})},
\label{eq:scl}
\end{equation}
\hl{where $y_i$ denotes the instance label and $\mathbf{u}_{y_i}$ is the prototype embedding of instance $y$}, computed as the mean of embeddings assigned to that instance. 
The same formulation in Eq.~\ref{eq:scl} is used in all three settings, differing only in how instance labels are obtained.

The final spatial contrastive learning loss is defined as a weighted combination of the three terms:
\begin{equation}
\mathcal{L}_{\text{SCL}} =
\lambda_{\text{iv}} \mathcal{L}_{\text{sc}}^{\text{iv}} +
\lambda_{\text{cv}} \mathcal{L}_{\text{sc}}^{\text{cv}} +
\lambda_{\text{3D}} \mathcal{L}_{\text{sc}}^{\text{3D}},
\end{equation}
\hl{where $\lambda_{\text{iv}}$, $\lambda_{\text{cv}}$, and $\lambda_{\text{3D}}$ balance the contributions of intra-view, cross-view, and 3D-space supervision.}

By enforcing instance consistency directly in the shared 3D feature space, spatial contrastive learning enables robust multi-person separation across views and provides instance-aware 3D tokens for subsequent SMPL regression.

In addition to instance-level contrastive learning, we further introduce a semantic contrastive objective on $P_v(\mathbf{u}_{i})$, $A_v(\mathbf{u}_{i})$, and $F_v(\mathbf{u}_{i})$ to improve the consistency of semantic representations across views.
The key observation is that, for a given person, semantic features extracted from different viewpoints should remain consistent, even under occlusion or partial visibility.
Such consistency is essential for learning a coherent 3D feature space that captures both spatial structure and semantic meaning.

Concretely, we treat the refined 3D features as semantic descriptors of the underlying 3D tokens.
For each human-related token, its 3D location is projected into multiple camera views, and only valid projections are selected based on visibility.
Semantic features sampled from different views but corresponding to the same 3D token and the same identity are encouraged to be similar, while features associated with different identities are pushed apart. The contrastive loss is the same as Eq.~\eqref{eq:scl}. 
This identity-aware contrastive supervision is applied directly to semantic features, rather than instance embeddings.

By enforcing semantic consistency across views at the 3D level, the proposed objective improves the model’s ability to extract stable semantic representations over the entire 3D space.
As a result, the learned features exhibit stronger view invariance and spatial coherence, which benefits subsequent SMPL regression.

\subsubsection{Training with Spatial Contrastive Learning}

During the contrastive learning stage, Gaussian parameters and the 3D feature tokens continue to be optimized jointly.

The spatial contrastive learning loss $\mathcal{L}_{\text{SCL}}$ is applied to both instance embeddings and semantic features sampled via multi-view reprojection.
Instance-level contrastive supervision encourages 3D tokens belonging to the same person to share similar embeddings, while separating tokens from different individuals.
In parallel, semantic contrastive supervision enforces view-consistent semantic representations for the same identity across different viewpoints.

Photometric reconstruction loss $\mathcal{L}_{\text{rgb}}$ is retained throughout this stage to preserve geometric consistency and prevent drift in the underlying 3D structure.
By jointly optimizing geometry, appearance, and contrastive objectives, the model learns a discriminative and spatially coherent 3D feature space that supports robust multi-person reasoning.

\subsection{SMPL Parameter Regression from Human Tokens}
\label{sec:smpl_regression}

With the refined 3D representation, we then recover structured human body models for each person in the scene.
Instead of iterative fitting, we adopt a feed-forward regression formulation that predicts SMPL parameters directly from the learned 3D representation.

\subsubsection{Person-token pooling with human-guided weighting}
Let $\{(\mathbf{p}_i, \tilde{\mathbf{f}}_i)\}_{i=1}^{N}$ denote the fused 3D tokens, where $\mathbf{p}_i\in\mathbb{R}^3$ is the 3D location and $\tilde{\mathbf{f}}_i\in\mathbb{R}^{C}$ is the refined feature (\secref{sec:cross_modal_fusion}).
We also obtain a human confidence score $h_i\in[0,1]$ for each token and a pseudo instance label $\ell_i\in\{0,1,\dots,P\}$, where $\ell_i=0$ indicates background and $\ell_i=p$ indicates the $p$-th person instance.

We first construct a point descriptor by concatenating 3D features and coordinates, followed by a projection MLP:
\begin{equation}
\mathbf{d}_i = \psi\big([\tilde{\mathbf{f}}_i;\mathbf{p}_i]\big)\in\mathbb{R}^{T},
\end{equation}
where $\psi(\cdot)$ denotes a lightweight MLP and $T$ is the token dimension.
We then pool a per-person token by confidence-weighted averaging over all tokens associated with the same person:
\begin{equation}
\mathbf{t}_p = 
\frac{
\sum_{i=1}^{N} \mathbb{I}[\ell_i=p]\; w_i\; \mathbf{d}_i
}{
\sum_{i=1}^{N} \mathbb{I}[\ell_i=p]\; w_i + \epsilon
},
\qquad
w_i = h_i^{\gamma},
\label{eq:person_token_pool}
\end{equation}
where $\gamma$ is a hyper-parameter controlling how strongly we emphasize high-confidence human tokens, and $\epsilon$ is a small constant.
This design effectively suppresses noisy or floating points by down-weighting low-confidence tokens.

We additionally keep the accumulated weight
\begin{equation}
s_p = \sum_{i=1}^{N} \mathbb{I}[\ell_i=p]\; w_i,
\end{equation}
which serves as a validity indicator for whether person $p$ is reliably observed and can be used to mask invalid instances during training.

\subsubsection{Feed-forward SMPL parameter regression}
Given the pooled person token $\mathbf{t}_p$, we regress SMPL parameters with a small MLP head:
\begin{equation}
\mathbf{z}_p = \rho(\mathbf{t}_p),
\end{equation}
followed by linear predictors for pose, shape, and translation:
\begin{equation}
\hat{\boldsymbol{\theta}}_p = \mathbf{W}_{\theta}\mathbf{z}_p,\quad
\hat{\boldsymbol{\beta}}_p = \mathbf{W}_{\beta}\mathbf{z}_p,\quad
\hat{\mathbf{t}}_p = \mathbf{W}_{t}\mathbf{z}_p,
\label{eq:smpl_regress}
\end{equation}
where $\hat{\boldsymbol{\theta}}_p\in\mathbb{R}^{144}$ denotes the 6D pose parameters, $\hat{\boldsymbol{\beta}}_p\in\mathbb{R}^{10}$ denotes the shape coefficients, and $\hat{\mathbf{t}}_p\in\mathbb{R}^{3}$ denotes the global translation.
Optionally, we also predict a global scale parameter $\hat{s}_p$ (implemented as a positive scalar via $softplus$) to compensate for scale ambiguity in monocular or normalized coordinate systems:
\begin{equation}
\hat{s}_p = \mathrm{softplus}(\mathbf{W}_{s}\mathbf{z}_p) + \epsilon.
\end{equation}

The regression outputs define an SMPL mesh for each person instance, enabling downstream tasks such as multi-view rendering, silhouette supervision, and instance-level evaluation without iterative optimization.

\subsubsection{Training with SMPL Regression}

In the final stage, we train the SMPL regression head while continuing to optimize the Gaussian representation with photometric supervision.
Given the pooled person tokens (\secref{sec:smpl_regression}), the network predicts SMPL pose, shape, and translation parameters for each person instance.

The SMPL regression is supervised using a combination of 2D reprojection, 3D joint alignment, and silhouette consistency:
\begin{equation}
\mathcal{L}_{\text{smpl}}
=
\lambda_{\text{j2d}}\mathcal{L}_{\text{j2d}}
+
\lambda_{\text{j3d}}\mathcal{L}_{\text{j3d}}
+
\lambda_{\text{seg}}\mathcal{L}_{\text{seg}}.
\end{equation}

The 2D reprojection loss $\mathcal{L}_{\text{j2d}}$ penalizes the discrepancy between projected SMPL joints and ground-truth 2D keypoints across views.
When available, the 3D joint loss $\mathcal{L}_{\text{j3d}}$ directly supervises predicted SMPL joints in 3D space.
In addition, a silhouette loss $\mathcal{L}_{\text{seg}}$ is applied by rendering the predicted SMPL meshes into each view and matching them with the corresponding human masks, which helps constrain body extent and resolve depth ambiguities under occlusions.

During this stage, photometric supervision $\mathcal{L}_{\text{rgb}}$ is retained to preserve geometric and appearance consistency of the Gaussian representation.
This joint optimization enables accurate recovery of structured SMPL parameters from the instance-aware 3D feature space.

\section{Experiments}\label{sec:Experiments}

We evaluate the proposed framework on challenging multi-view human reconstruction benchmarks under unconstrained conditions.
Our experiments are designed to validate the effectiveness of maintaining a unified human-aware 3D feature space, as well as the contributions of reprojection-based enhancement, spatial contrastive learning, and 3D token–based SMPL regression.

\subsection{Datasets}
\label{datasets}
\textbf{EgoHumans}~\cite{egohumans} is a multi-camera, multi-person dataset designed for evaluating human pose estimation and reconstruction in real-world environments.
It consists of video sequences captured by multiple synchronized cameras, where two to four individuals perform everyday activities.
The dataset covers a wide range of scene scales and layouts, including both indoor and outdoor environments, and exhibits frequent inter-person occlusions, viewpoint variations, and camera placement irregularities.
These characteristics make EgoHumans particularly suitable for evaluating the robustness of multi-view reconstruction methods under realistic conditions.
Following common practice, we construct train/test splits at the scene level: for each scene, one sequence is randomly selected for testing, while the remaining sequences are used for training.

\textbf{OcMotion}~\cite{huang2022occluded} is a multi-camera dataset focusing on single-person motion under severe object-induced occlusions.
It contains diverse everyday activities where the human body is partially or heavily occluded by surrounding objects, resulting in incomplete and view-dependent observations.
OcMotion provides a complementary evaluation setting that emphasizes robustness to dynamic occlusion and missing visual evidence.
For OcMotion, we follow the official dataset split as defined in the original benchmark.

\subsection{Metrics}
\label{sec:metrics}

We evaluate both human reconstruction accuracy and camera estimation quality using standard metrics.

\hl{For human reconstruction, we report \textbf{CA-MPJPE} (Camera-aligned MPJPE, meters) first aligns the estimated camera parameters with the ground-truth camera parameters, and then computes MPJPE between the predicted and ground-truth 3D joints.
We additionally report \textbf{GA-MPJPE} (Procrustes-aligned MPJPE, meters), which aligns the predicted multi-person configuration with the ground-truth multi-person layout as a whole, and evaluates the relative spatial relationships among all individuals. 
We further report \textbf{PA-MPJPE} (Group-aligned MPJPE, meters), which independently applies Procrustes alignment to each predicted person and evaluates per-person pose accuracy for articulated human reconstruction.
These metrics together provide a more comprehensive and interpretable evaluation of both local pose accuracy and global multi-person spatial structure.}

For camera evaluation, we report \textbf{AE} (angular error) and \textbf{s-TE} (scale-normalized translation error) to measure rotation and translation accuracy, respectively.
We additionally report \textbf{s-CCA@10} and \textbf{AUC@10}, which evaluate camera consistency within a fixed angular threshold.
Lower AE and s-TE, and higher s-CCA@10 and AUC@10 indicate more accurate and consistent camera estimation.

\begin{table}[ht]
\centering
\caption{
Quantitative comparison of multi-view human reconstruction accuracy on OcMotion.
}
\label{tab:cmp_ocmotion}
\begin{tabular}{lcc}
\hline
\textbf{Method} & \textbf{CA-MPJPE} $\downarrow$ & \textbf{PA-MPJPE} $\downarrow$ \\
\hline
Multi-HMR~\cite{baradel2024multi} & - & 0.06 \\
DMMR~\cite{huang2023dmmr} & - & \textbf{0.04} \\
HSfM~\cite{muller2025reconstructing} & 0.35 & 0.05 \\
Ours & \textbf{0.19} & \textbf{0.04} \\
\hline
\end{tabular}
\end{table}

\begin{table*}[t]
\centering
\caption{
Quantitative comparison of multi-view human reconstruction accuracy on EgoHumans under non-severely and severely occluded settings.
}
\label{tab:cmp_egohumans}
\resizebox{1.0\textwidth}{!}{
\begin{tabular}{lccc|ccc}
\hline
\multirow{2}{*}{\textbf{Method}} 
& \multicolumn{3}{c|}{\textbf{Non-severely occluded}} 
& \multicolumn{3}{c}{\textbf{Severely occluded}} \\
& CA-MPJPE$\downarrow$ & GA-MPJPE$\downarrow$ & PA-MPJPE$\downarrow$
& CA-MPJPE$\downarrow$ & GA-MPJPE$\downarrow$ & PA-MPJPE$\downarrow$ \\
\hline
Multi-HMR~\cite{baradel2024multi} & - & 0.85 & 0.10 & - & 1.23 & \textbf{0.12} \\
DMMR~\cite{huang2023dmmr} & - & 0.54 & 0.23 & - & 0.81 & 0.20 \\
HSfM~\cite{muller2025reconstructing} & 0.84 & 0.34 & 0.08 & 1.18 & 0.53 & 0.13 \\
Ours & \textbf{0.80} & \textbf{0.21} & \textbf{0.07} & \textbf{0.89} & \textbf{0.28} & 0.14 \\
\hline
\end{tabular}
}
\end{table*}

\subsection{Results}\label{sec:Comparison}

We compare our method with representative state-of-the-art multi-view reconstruction approaches on the EgoHumans {\cite{egohumans}} and OcMotion {\cite{huang2022occluded}} dataset.
In particular, we include comparison with HSfM~\cite{muller2025reconstructing}, a recent framework for jointly reconstructing people, scene structure, and camera parameters from multi-view images. 
HSfM represents a strong baseline that integrates geometry and human modeling in a unified optimization framework.

Tables~\ref{tab:cmp_ocmotion} and~\ref{tab:cmp_egohumans} report quantitative results on OcMotion and EgoHumans under different occlusion settings, respectively, while Table~\ref{tab:cmp2} presents camera pose estimation accuracy.

Table~\ref{tab:cmp_ocmotion} \hl{reports results on OcMotion, which involves challenging single-person scenarios with severe object-induced occlusions. Our method achieves the best CA-MPJPE and competitive PA-MPJPE, outperforming Multi-HMR, DMMR and HSfM in global reconstruction accuracy. }
Table~\ref{tab:cmp_egohumans} \hl{further evaluates performance on the EgoHumans dataset under non-severely and severely occluded scenarios. In particular, we define severe occlusion as the presence of a subject occluded in more than 50\% of the views. All remaining samples are grouped into the non-severely occluded subset. Across all settings, our method consistently achieves strong CA-MPJPE and GA-MPJPE, indicating robust multi-person spatial reconstruction under complex occlusions. Compared to Multi-HMR and DMMR, our method better preserves global spatial consistency, especially under heavy occlusion where cross-view correspondence becomes highly ambiguous. HSfM shows competitive inter-person structure modeling but suffers from larger absolute reconstruction errors. Overall, our method achieves consistently strong global reconstruction accuracy while maintaining competitive or superior local pose accuracy across both datasets and occlusion levels.}

As shown in Table~\ref{tab:cmp2}, our method consistently achieves substantially lower camera pose errors across all metrics on both EgoHumans and OcMotion.
In particular, the improvements in AE and s-TE indicate accurate estimation of absolute camera orientation and translation, while the high s-CCA@10 and AUC@10 scores demonstrate strong cross-view consistency.
These results suggest that maintaining a unified 3D feature space enables reliable joint reasoning over camera geometry and scene structure without explicit calibration.

\begin{figure*}[!t]
\centering
\includegraphics[width=\linewidth]{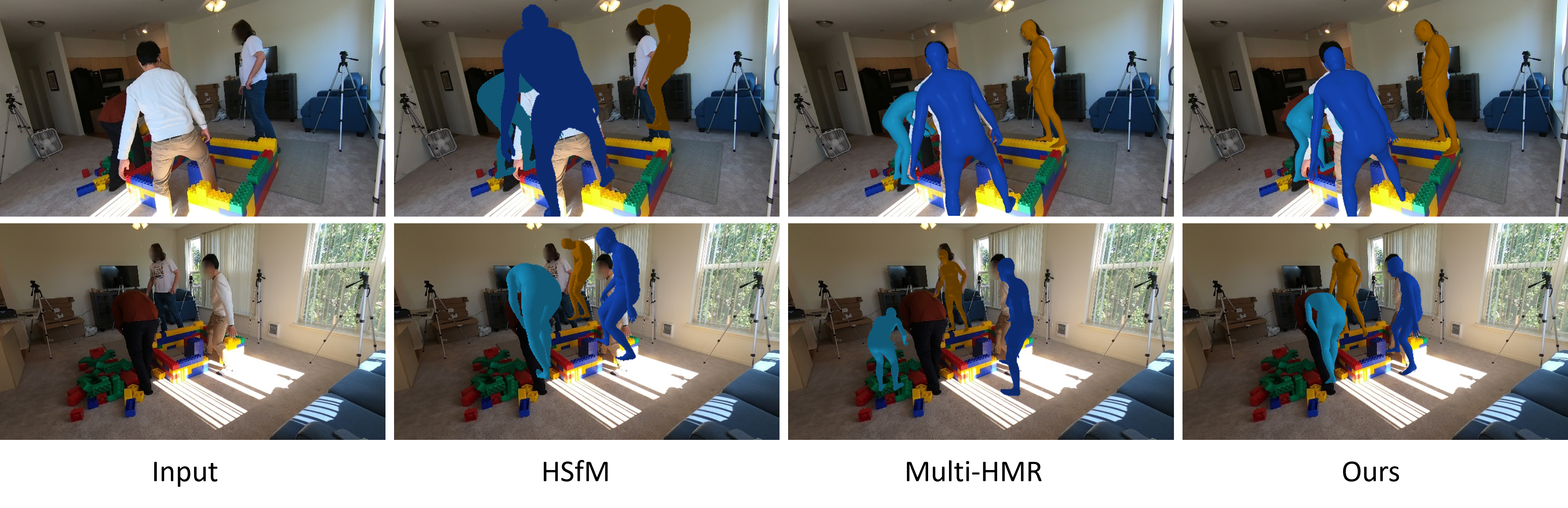}
\caption{
Qualitative Comparison of Multi-View Human Reconstruction.
}
\label{fig:result2}
\end{figure*}

Figure~\ref{fig:result2} presents qualitative reprojection comparisons.
HSfM tends to prioritize relative spatial relationships between individuals, which often leads to inaccurate reprojection results and visible misalignments in the image plane.
Multi-HMR produces accurate reprojections in the input views but lacks explicit 3D reasoning, resulting in inconsistent reconstructions across different viewpoints.
In comparison, our method produces accurate and consistent reprojections across views, reflecting improved multi-view coherence and pose estimation.

Finally, we visualize the reconstructed 3D scene together with the recovered SMPL bodies in Figure~\ref{fig:result1}.
The results show that our framework is able to reconstruct both scene geometry and human bodies in a coherent and physically plausible manner, further validating the effectiveness of maintaining a unified human-aware 3D feature space.


\begin{table*}[t]
\begin{center}
\caption{
Quantitative comparison of Camera pose accuracy on EgoHumans and OcMotion.
}
\label{tab:cmp2}
\resizebox{1.0\textwidth}{!}{
\begin{tabular}{lcccc|cccc}
\hline
\multirow{2}{*}{\textbf{Method}} & \multicolumn{4}{c}{\textbf{EgoHumans}} & \multicolumn{4}{c}{OcMotion}\\ 
& AE$\downarrow$
& s-TE$\downarrow$
& s-CCA@10$\uparrow$
& AUC@10$\uparrow$
& AE$\downarrow$
& s-TE$\downarrow$
& s-CCA@10$\uparrow$ 
& AUC@10$\uparrow$\\ 
\hline
UnCaliPose\cite{xu2022uncalibrated} & 130.68 & 4.38 & 0.02 &0.23 &154.43 &4.31 &0.01 &0.19\\
HSfM\cite{muller2025reconstructing} &29.58  &0.77  &0.17  & 0.62 & 0.97 &0.13 &0.50 &0.80\\
Ours & \textbf{0.60} & \textbf{0.03} & \textbf{0.95} & \textbf{0.94} &\textbf{0.36} &\textbf{0.04} &\textbf{0.99} &\textbf{0.94}\\ 
\hline
\end{tabular}
}
\end{center}
\end{table*}

\begin{figure*}[!t]
\centering
\label{fig:result1}
\includegraphics[width=\linewidth]{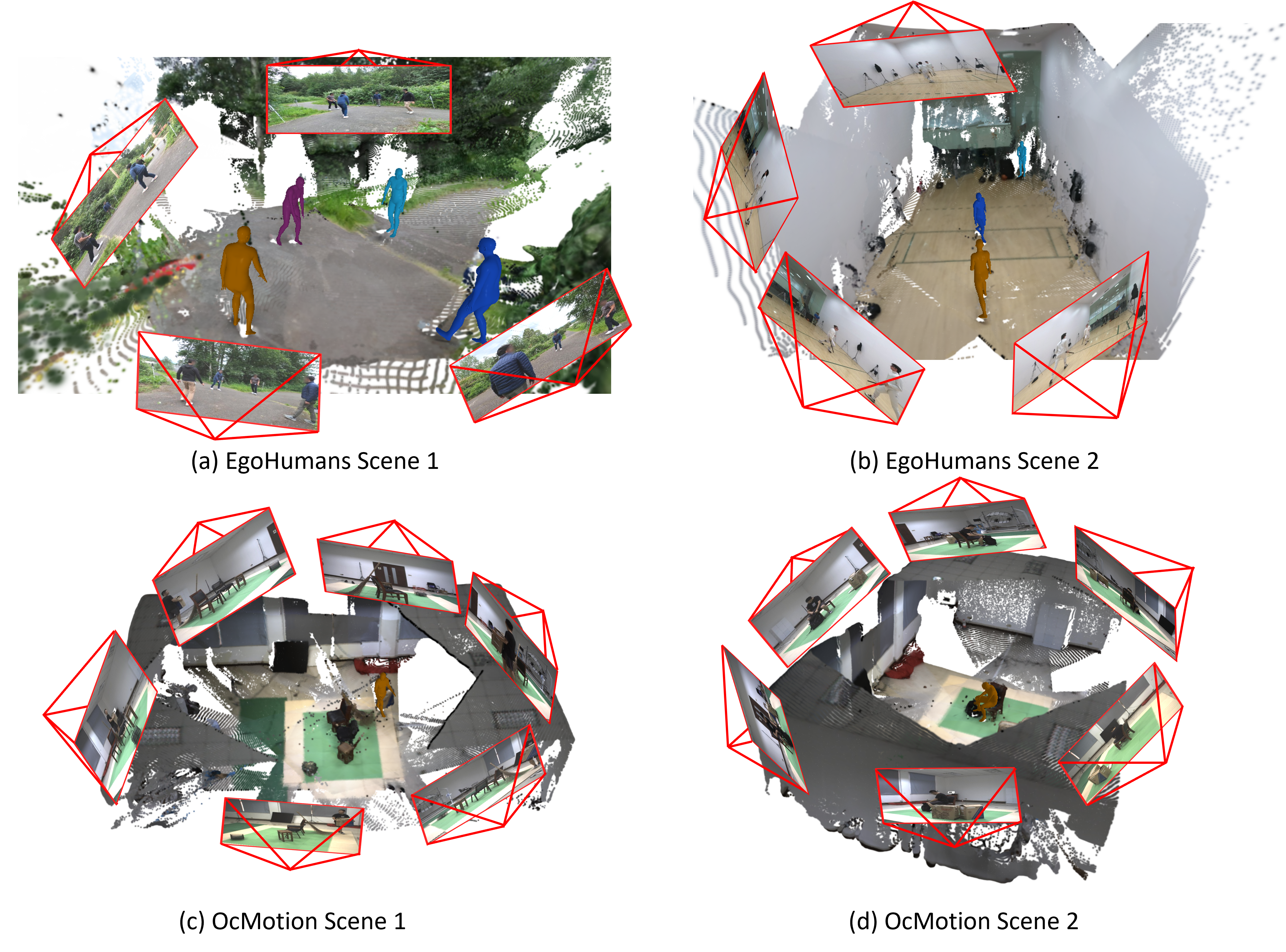}
\caption{
Qualitative results of the proposed method on EgoHumans and OcMotion.
We visualize the reconstructed 3D scene together with the recovered SMPL bodies from multiple views.
The results demonstrate that our method produces spatially coherent human reconstructions.
}
\label{fig:result1}
\end{figure*}

\begin{table*}[t]
\begin{center}
\caption{
Runtime breakdown (s) of the proposed method under different numbers of input frames.
}
\label{tab:runtime}
\begin{tabular}{lcccccc}
\hline
\textbf{Input Frames} & 1 & 2 & 4 & 8 & 16 & 32 \\
\hline
3D Scene Reconstruction & 0.099 & 0.131 & 0.219 & 0.382 & 0.812 & 1.886 \\
Instance-level Human Segmentation & 0.064 & 0.184 & 0.471 & 0.501 & 0.501 & 0.488 \\
SMPL Regression & 0.046 & 0.046 & 0.048 & 0.052 & 0.084 & 0.180 \\
\hline
Total & 0.209 & 0.361 & 0.729 & 0.936 & 1.397 & 2.554 \\
\hline
\end{tabular}
\end{center}
\end{table*}
\subsection{Runtime and Memory Analysis}
\hl{
We further analyze the computational efficiency of the proposed framework under different numbers of input views, and compare it with a representative optimization-based baseline HSfM.

Since all components in our framework are implemented in a feed-forward regression manner, the overall computational cost is primarily determined by the number of input frames rather than iterative optimization or test-time refinement. This makes the proposed method fundamentally different from optimization-based reconstruction pipelines.

Table}~\ref{tab:runtime} \hl{reports the runtime breakdown of different stages in our pipeline, including 3D scene reconstruction, instance-level human segmentation, and SMPL regression. We observe that the 3D scene reconstruction stage scales with the number of input views due to additional depth estimation and feature lifting from VGGT. The instance-level human segmentation stage depends on the number of reconstructed human-related 3D points and gradually stabilizes as spatial density converges. The SMPL regression module remains lightweight and relatively stable across different input sizes.

Overall, although runtime increases with more input views, the growth is sub-linear rather than strictly proportional, indicating good scalability in multi-view settings. In contrast, optimization-based methods such as HSfM require iterative refinement, leading to significantly higher computational cost as shown in Table}~\ref{tab:cmp_runtime}.

\hl{
We further report a comparison of runtime and GPU memory consumption between our method and HSfM. As shown in Table}~\ref{tab:cmp_runtime}\hl{, our approach achieves substantially lower runtime while maintaining comparable or higher memory efficiency under increasing numbers of input views. This demonstrates that the proposed framework provides a favorable trade-off between reconstruction accuracy and computational efficiency in multi-view scenarios.
}
\begin{table}[t]
\caption{
Runtime and peak GPU memory comparison with HSfM under different numbers of input frames.
}
\label{tab:cmp_runtime}
\begin{tabular}{lccccc}
\hline
\textbf{Method} & \textbf{Metric} & 1 & 2 & 4 & 8 \\
\hline
HSfM & Time (s) & 170 & 177 & 199 & 220 \\
HSfM & Mem. (GB) & 4.62 & 5.13 & 6.29 & 10.79 \\
Ours & Time (s) & 0.209 & 0.361 & 0.729 & 0.936 \\
Ours & Mem. (GB) & 8.26 & 8.72 & 11.20 & 15.82 \\
\hline
\end{tabular}
\end{table}

\subsection{Cross-view Association Evaluation}

\hl{To further evaluate the quality of cross-view instance association, we further apply the AIDP}~\cite{gan2021self} \hl{metrics for cross-view association evaluation. AIDP evaluates the precision of subject association across views by computing pairwise identity matching accuracy over all view pairs and averaging the results. This metric directly reflects the consistency of instance-level correspondence in multi-view settings.

We first compare our method with HSfM under this evaluation protocol. In addition, to further examine whether geometric cues alone are sufficient for reliable association, we construct a VGGT-based Pose-Aware ReID baseline. Specifically, given detected human bounding boxes, we extract keypoint predictions and confidence scores from VGGT, and perform confidence-weighted aggregation to obtain compact pose-centric descriptors. Cross-view identity association is then performed via nearest-neighbor matching based on Euclidean distance.

As shown in Table}~\ref{tab:aidp}\hl{, our method achieves the highest AIDP score, outperforming both HSfM and the VGGT-based baseline. This indicates that geometry- or pose-based heuristics can provide reasonable cues for association, but remain limited under challenging multi-view conditions. In contrast, our spatial contrastive learning explicitly enforces instance-level consistency in the unified 3D feature space, leading to more robust cross-view identity alignment.}

\begin{table}[t]
\centering
\caption{Cross-view instance association evaluation using AIDP metric. Higher is better.}
\label{tab:aidp}
\begin{tabular}{lc}
\hline
\textbf{Method} & \textbf{AIDP} $\uparrow$ \\
\hline
HSfM~\cite{muller2025reconstructing} & 75.68 \\
VGGT-based Pose-Aware ReID & 82.02 \\
Ours & \textbf{98.54} \\
\hline
\end{tabular}
\end{table}

\subsection{Ablation Study}
\label{sec:Ablation}
\subsubsection{Architecture Ablation}
We conduct ablation studies on EgoHumans to analyze the contributions of key components in the proposed framework.
In particular, we investigate how different design choices affect multi-view human perception, instance consistency, and SMPL regression from the unified 3D feature space.

Specifically, we consider the following variants.
\textbf{w/o unified 3D feature space} removes explicit 3D perception and directly regresses human bodies from per-view 2D feature maps.
\textbf{w/o reprojection enhancement} disables reprojection-based pose feature sampling and regresses SMPL parameters solely from the initial 3D feature space containing geometric cues.
\textbf{w/o SCL} removes spatial contrastive learning and aggregates features based on 2D instance segmentation without enforcing instance consistency in 3D.

Table~\ref{tab:ablation} summarizes the ablation results.
Removing the unified 3D feature space leads to a significant performance drop.
Without explicit 3D perception, the model relies solely on view-dependent 2D features, which are highly sensitive to occlusion and viewpoint variation.
As a result, it becomes difficult to associate observations of the same individual across views, leading to degraded reconstruction accuracy.

Disabling reprojection-based enhancement also degrades performance.
In this setting, SMPL parameters are regressed only from the initial 3D feature space that mainly encodes geometric information.
Without injecting pose-related image features via reprojection, optimization becomes more difficult and convergence is less stable, resulting in inferior reconstruction quality.

Removing spatial contrastive learning further impacts performance.
When instance consistency is enforced only through 2D instance segmentation, feature aggregation becomes vulnerable to segmentation noise and view-dependent errors.
This leads to contaminated 3D features and reduced robustness, especially in the presence of occlusion and clutter.

\begin{table*}[t]
\begin{center}
\caption{
Ablation study on key components of the proposed framework on EgoHumans.
}
\label{tab:ablation}
\begin{tabular}{lccc}
\hline
\textbf{Method} & CA-MPJPE$\downarrow$ & GA-MPJPE$\downarrow$ & PA-MPJPE$\downarrow$\\
\hline
    w/o unified 3D feature space &3.12 &3.28 &2.53\\
    w/o reprojection enhancement &1.28 &1.42 &1.74\\
    w/o SCL &1.94 &0.49 &0.37\\
    Ours &0.95 &0.30 &0.18\\
\hline
\end{tabular}
\end{center}
\end{table*}

\subsubsection{Robustness to Geometry Initialization}

\hl{To further evaluate the robustness of our framework to geometry initialization, we replace the VGGT backbone with AnySplat}~\cite{jiang2025anysplat}\hl{, a representative geometry-aware reconstruction method that provides depth and camera priors with Gaussian Splatting-based refinement.

Both VGGT and AnySplat are geometry-aware reconstruction pipelines that produce compatible initializations of depth, features, and camera parameters, making them suitable for evaluating the sensitivity of our framework to different geometry backbones.
Unlike VGGT, which directly predicts geometry and feature representations, AnySplat introduces an additional rendering-based refinement stage, resulting in slightly different camera and point cloud quality. Nevertheless, both methods share a similar underlying geometry foundation, allowing a fair comparison of backbone robustness.

As shown in Table}~\ref{tab:backbone}\hl{, the performance remains highly consistent under both initialization strategies, with only marginal differences across all evaluation metrics. This indicates that our method is largely robust to the choice of geometry initialization. The results further suggest that our method is robust to the choice of geometry initialization, and benefits primarily from the unified 3D feature representation and spatial contrastive learning.}

\begin{table*}[t]
\begin{center}
\caption{
Robustness to geometry initialization.
We compare VGGT-based initialization with an alternative AnySplat-based initialization.
}
\label{tab:backbone}
\begin{tabular}{lccc}
\hline
\textbf{Geometry Initialization} & CA-MPJPE$\downarrow$ & GA-MPJPE$\downarrow$ & PA-MPJPE$\downarrow$\\
\hline
w/ AnySplat~\cite{jiang2025anysplat} & 1.02 & 0.37 & 0.19\\
VGGT (default) & 0.95 & 0.30 & 0.18\\
\hline
\end{tabular}
\end{center}
\end{table*}

\subsubsection{Analysis of 3D Human Awareness}
The limitations of relying solely on 2D instance segmentation further motivate a deeper analysis of human perception in 3D.
Although 2D segmentation-based aggregation can achieve high coverage and include most human-related features, it exhibits structural weaknesses when applied to multi-view 3D reconstruction.

In practice, segmentation errors near object boundaries and occlusion regions are common.
When such masks are lifted into 3D, these inaccuracies lead to enlarged selection regions and ray-like leakage into the background, contaminating the aggregated 3D features.
Moreover, 2D instance segmentation does not naturally provide consistent instance identities across views.
Associating masks belonging to the same person from different viewpoints requires additional cross-view matching, which is highly sensitive to occlusion, truncation, and segmentation noise.
Inconsistent instance assignments across views further degrade feature aggregation and undermine multi-view coherence.

These observations indicate that accurate human perception in multi-view settings cannot be reliably achieved by inheriting instance information from 2D segmentation alone.
Instead, human awareness and instance consistency should be learned directly in the shared 3D feature space, where multi-view observations can be aggregated and regularized in a geometry-aware manner.
In the following, we further validate this design choice.

On a single-person setting with per-view masks and ground-truth SMPL models, we compare the proposed learned 3D human confidence with a non-learned baseline that fuses 2D masks into 3D via visibility-aware averaging.
Ground-truth human labels for 3D tokens are derived based on their distance to the SMPL surface.

Table~\ref{tab:human_conf} reports token-level Precision, Recall, F1 score, and PR-AUC.
The 2D mask fusion baseline achieves very high recall but suffers from extremely low precision, indicating severe background leakage.
In contrast, the learned 3D confidence substantially improves precision and F1 score while maintaining competitive recall, and achieves a significantly higher PR-AUC, demonstrating better global separability and calibration.

\begin{figure*}[t]
    \centering
    \includegraphics[width=\linewidth]{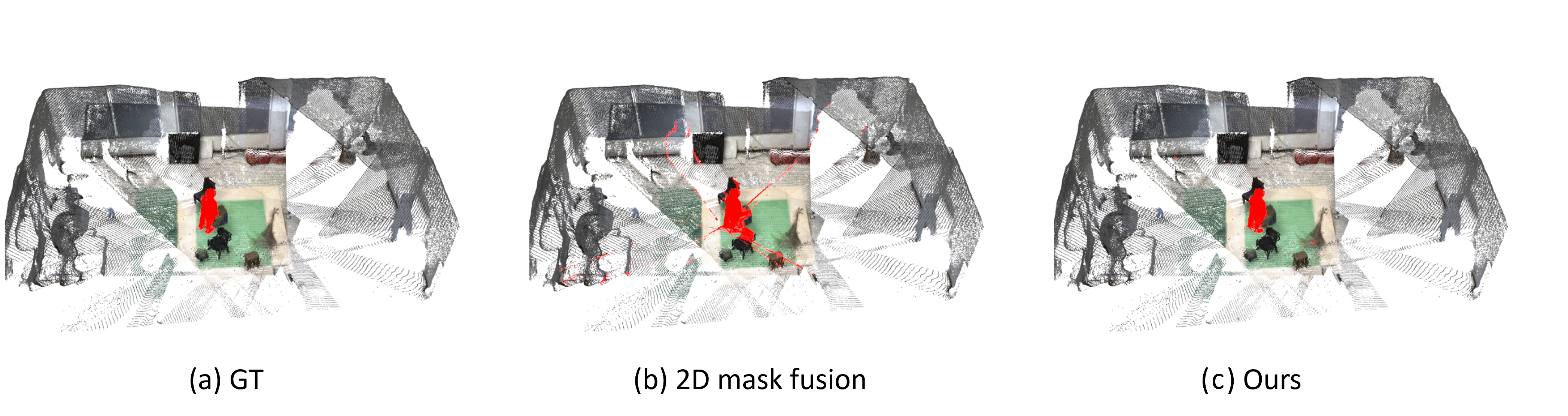}
    \caption{
    Qualitative comparison of 3D human confidence fields.
    While 2D mask fusion covers most human regions, it introduces substantial false positives in background and occluded areas.
    In contrast, the learned 3D confidence produces a spatially compact and clean human region with significantly reduced background noise.
    }
    \label{fig:human_conf_vis}
\end{figure*}

Figure~\ref{fig:human_conf_vis} provides qualitative evidence.
Directly lifting 2D masks into 3D introduces ray-like artifacts and background noise around occlusion boundaries, whereas the learned 3D confidence yields a spatially compact and coherent human region.
These results indicate that human awareness is not merely inherited from 2D segmentation, but emerges as a spatially consistent property learned in the unified 3D feature space, forming a reliable foundation for subsequent human-guided modules.

\begin{table}[t]
\centering
\caption{Comparison between learned 3D human confidence and 2D mask fusion on the single-person dataset.
Evaluation is performed at the token level using ground-truth SMPL supervision.}
\label{tab:human_conf}
\begin{tabular}{lcccc}
\toprule
Method & Precision $\uparrow$ & Recall $\uparrow$ & F1 $\uparrow$ & PR-AUC $\uparrow$ \\
\midrule
2D mask fusion & 0.420 & \textbf{0.989} & 0.590 & 0.006 \\
Ours (3D learned) & \textbf{0.886} & 0.825 & \textbf{0.855} & \textbf{0.863} \\
\bottomrule
\end{tabular}
\end{table}

\section{Conclusion}\label{sec:Conclusion}


In this paper, we presented a novel feed-forward framework for multi-view multi-person 3D reconstruction.
Departing from conventional pipelines that rely on view-dependent 2D reasoning or tightly coupled optimization, our method explicitly maintains a shared 3D representation where geometry, semantics, and human identity are jointly encoded and progressively refined.
By initializing the 3D space with a geometry-aware prior and 3D Gaussian representation, the framework establishes stable spatial anchors under unconstrained camera settings.
Building upon this foundation, spatial contrastive learning enhances the discriminability and consistency of the 3D features by leveraging multi-view semantic cues, enabling robust multi-person separation and cross-view association in the presence of occlusion and partial visibility.
Structured human body models are then recovered in a feed-forward manner directly from instance-aware 3D tokens, avoiding iterative fitting and allowing camera parameters and human representations to be jointly optimized.
Extensive experiments on challenging multi-view benchmarks demonstrate that the proposed approach achieves competitive or superior human reconstruction accuracy while significantly improving camera estimation quality.

\hl{While the present work focuses on static multi-view reconstruction, the proposed framework can be naturally extended to 4D dynamic settings by incorporating temporal association of 3D tokens across sequential frames. In particular, the instance-aware 3D representation provides a structured and identity-consistent basis that can naturally support temporal correspondence, enabling potential extensions to articulated motion and moving-camera scenarios. We further note that the current formulation does not explicitly model temporal evolution, and therefore future work may explore temporal update mechanisms such as token propagation, merging, and splitting within the same 3D Gaussian representation.}

\hl{We also acknowledge that our framework relies on VGGT as a geometry-aware foundation model for initializing depth, features, and camera parameters. While VGGT provides strong generalization for constructing the initial 3D feature space, its performance is not perfect and may be limited in cases of large viewpoint gaps, low-texture regions, or challenging depth discontinuities, which can affect downstream reconstruction quality. Nevertheless, our method is not tightly coupled to VGGT itself and operates on the constructed 3D feature representation. Therefore, it can be naturally integrated with alternative or improved geometry foundation models in future work.}

\hl{We believe that extending the framework toward full 4D reconstruction and improving robustness under more challenging geometry initialization remain important and promising directions for future research.}

\medskip
\noindent\textbf{Funding}  This work was supported in part by National Key R\&D Program of China (2023YFC3082100), the Open Project Program of State Key Laboratory of Virtual Reality Technology and Systems, Beihang University (No.VRLAB2026C04), and Science Fund for Distinguished Young Scholars of Tianjin (No.22JCJQJC00040).

\medskip
\noindent\textbf{Data availability} The datasets used in this study are publicly available. Detailed descriptions and corresponding references are provided in Section~\ref{datasets}.

{
    \small
    \bibliography{main}
}

\end{document}